\documentclass[letterpaper, 10 pt, conference]{ieeeconf}  
\usepackage{amsmath}
\usepackage{graphicx}
\IEEEoverridecommandlockouts                              

\usepackage{float} 
\usepackage{amsfonts}
\usepackage{url}
\usepackage{amssymb}   
\title{\LARGE \bf
Multi-Gait Learning for Humanoid Robots Using Reinforcement Learning with Selective Adversarial Motion Prior
}

\author{
\authorblockN{
Yuanye Wu$^{1*}$, 
Keyi Wang$^{1*}$, 
Linqi Ye$^{1}$,
and Boyang Xing$^{2}$
}
\thanks{*These authors contributed equally to this work.}
\thanks{$^{1}$Y. Wu, K. Wang, and L. Ye are with the School of Future Technology, Shanghai University, Shanghai, 200444, China (Corresponding author: Linqi Ye, email: yelinqi@shu.edu.cn)
}
\thanks{$^{2}$B. Xing is with National and Local Co-Built Humanoid Robotics Innovation Center, 201203 Shanghai, China}
}

\begin{document}

\maketitle
\thispagestyle{empty}
\pagestyle{empty}
\begin{abstract}
Learning diverse locomotion skills for humanoid robots in a unified reinforcement learning framework remains challenging due to the conflicting requirements of stability and dynamic expressiveness across different gaits. We present a multi-gait learning approach that enables a humanoid robot to master five distinct gaits---walking, goose-stepping, running, stair climbing, and jumping---using a consistent policy structure, action space, and reward formulation. The key contribution is a selective Adversarial Motion Prior (AMP) strategy: AMP is applied to periodic, stability-critical gaits (walking, goose-stepping, stair climbing) where it accelerates convergence and suppresses erratic behavior, while being deliberately omitted for highly dynamic gaits (running, jumping) where its regularization would over-constrain the motion. Policies are trained via PPO with domain randomization in simulation and deployed on a physical 12-DOF humanoid robot through zero-shot sim-to-real transfer. Quantitative comparisons demonstrate that selective AMP outperforms a uniform AMP policy across all five gaits, achieving faster convergence, lower tracking error, and higher success rates on stability-focused gaits without sacrificing the agility required for dynamic ones.
\end{abstract}

\section{INTRODUCTION}

Humanoid robots hold the promise of operating in environments designed for humans, but realizing this potential requires the ability to execute multiple locomotion behaviors adaptively~\cite{c1}. Practical deployment demands not only walking for efficient traversal but also running for rapid movement, jumping for gap crossing or obstacle negotiation, stair climbing for multi-level navigation, and specialized gaits such as goose-stepping for structured or ceremonial scenarios. Each of these gaits imposes fundamentally different requirements on the control system: walking emphasizes energy efficiency and steady balance, running requires rapid force generation and aerial phase management, jumping demands maximal vertical impulse in a short time window, stair climbing needs precise foot placement and clearance control, and goose-stepping requires sustained leg extension with strict posture constraints. The central challenge is not merely acquiring each gait in isolation but learning them within a unified framework that shares policy architecture, observation space, and reward structure.

Reinforcement learning has emerged as a compelling approach for learning complex motor skills without requiring explicit analytical models. By interacting with a physics simulator at scale, RL agents can discover control strategies that are difficult to hand-design, especially for underactuated, high-degree-of-freedom systems like humanoid robots~\cite{c2}. The key advantage of RL over traditional model-based control is its ability to handle nonlinear dynamics, contact switching, and actuator limits in a unified optimization process. Recent work has demonstrated impressive single-gait capabilities, including robust walking on flat and uneven ground, stair traversal with foot clearance, and basic disturbance recovery~\cite{c3,c4}. However, these studies typically train a separate policy for each task, and each policy is tuned with its own reward design and hyperparameter configuration. Extending RL to multiple gaits under a single framework introduces new difficulties: the policy must reconcile conflicting dynamics across behaviors, the reward function must remain effective across gaits with vastly different motion characteristics, and the observation space must encode sufficient information for all behaviors without becoming redundant or misleading for any individual gait~\cite{c5}.

Adversarial Motion Prior (AMP) addresses a key difficulty in RL-based locomotion---producing natural, stable motions that transfer well to hardware~\cite{c6}. By training a discriminator to distinguish between agent-generated and reference motion clips, AMP regularizes the policy toward demonstrated human-like behavior distributions. This has been shown to reduce joint jitter, improve foot placement consistency, and produce smoother trajectories that are more compatible with real actuator dynamics~\cite{c7,c8}. However, existing work that employs AMP does so uniformly: once an AMP module is introduced, it remains active for all behaviors throughout training. This raises an important but underexplored question: \emph{is AMP always beneficial, or can it be counterproductive for certain gait types?}

Our observation is that the effectiveness of AMP depends critically on the nature of the target gait. AMP excels at reinforcing periodic, repetitive motion patterns with moderate joint amplitude---exactly the characteristics of walking, goose-stepping, and stair climbing. For these gaits, the reference motion distribution closely matches the desired policy output, and the adversarial reward provides a stabilizing inductive bias that accelerates convergence and suppresses erratic behavior. Conversely, for gaits requiring large joint excursions, rapid acceleration, and explosive dynamics---such as running and jumping---the available human motion references are limited, and the AMP discriminator's regularization can act as an undesired constraint. In these cases, the policy needs to explore beyond the reference manifold to discover strategies that fully utilize the robot's actuator capacity, and the AMP penalty can prevent the agent from reaching the necessary motion extremes.

This mismatch motivates a selective approach: rather than applying AMP universally or discarding it entirely, we propose to use it where it helps and disable it where it hinders. This strategy requires understanding which gait characteristics align with the AMP assumption and which deviate from it, and making an informed decision for each locomotion mode.


The contributions of this paper are as follows:

\begin{enumerate}
    \item \textbf{A unified RL framework} that learns five practical gaits for humanoid robots---walking, goose-stepping, running, jumping, and stair climbing---using a consistent policy architecture and reward function based on sinusoidal joint reference tracking. All gaits share the same observation space, network structure, and reward formulation, differing only in reference trajectory parameters and reward weights.
    \item \textbf{A selective Adversarial Motion Prior strategy} that applies AMP to periodic, stability-critical gaits while omitting it for highly dynamic, agile gaits. We provide empirical analysis demonstrating that AMP improves convergence speed and tracking accuracy for walking-type gaits but restricts motion amplitude and expressiveness for running and jumping.
    \item \textbf{Full real-robot validation} of all five gaits with quantitative comparisons including tracking error, convergence speed, success/fall rate, and AMP versus non-AMP performance. We deploy every learned policy on a physical humanoid platform, providing a reproducible real-robot baseline for multi-gait learning.
\end{enumerate}
\section{Methodology}
\subsection{Overview}
\begin{figure}[t]
    \centering
    \includegraphics[width=0.46\textwidth]{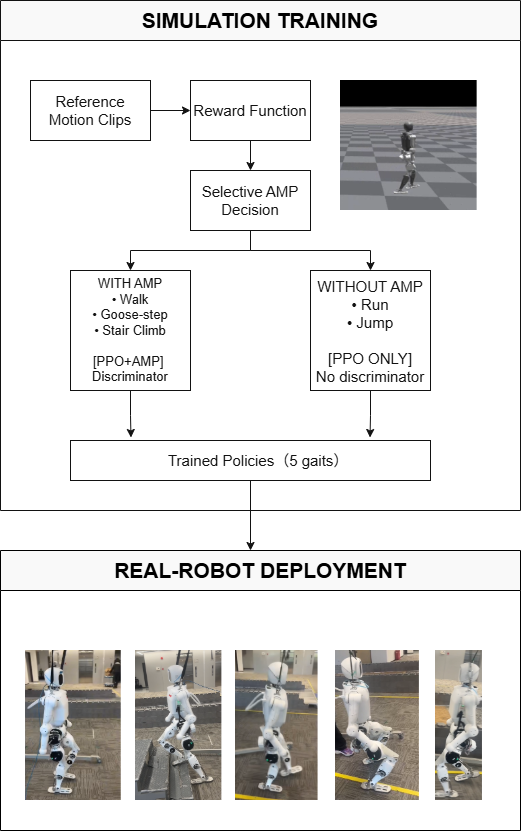}
    \caption{Overview of the multi-gait learning pipeline. Policies are trained in simulation with or without AMP depending on gait type, then deployed on the physical robot.}
    \label{fig:overview}
\end{figure}

We propose a unified reinforcement learning framework that learns five distinct locomotion gaits for a 12-DOF humanoid robot: walking, goose-stepping, running, jumping, and stair climbing. The framework consists of three main components. First, a unified RL formulation with a consistent observation space, action space, and reward structure is used to train all five gaits, where each gait is distinguished only by its sinusoidal reference trajectory parameters and reward weight configuration. Second, a selective AMP strategy is applied: AMP is enabled for periodic, stability-critical gaits (walking, goose-stepping, stair climbing) and disabled for highly dynamic gaits (running, jumping). Third, policies trained in simulation with extensive domain randomization are deployed directly on a physical humanoid robot without fine-tuning. Figure~\ref{fig:overview} illustrates the overall pipeline.

\subsection{Reinforcement Learning Formulation}

We formulate each gait learning task as a Markov Decision Process (MDP) defined by the tuple $(\mathcal{S}, \mathcal{A}, r, \gamma, \rho_0)$. All five gaits share the same MDP structure, differing only in the reference trajectory generation, reward weight parameters, and AMP configuration.

\subsubsection{State Space}

The observation space received by the policy at each timestep $t$ consists of proprioceptive states, reference signals, and command inputs. The per-frame observation vector $o_t \in \mathbb{R}^{50}$ is constructed as the concatenation of the following components:

\begin{itemize}
    \item \textbf{Base linear velocity} $v_b \in \mathbb{R}^3$: expressed in the robot's body frame, scaled by a factor of 2.0.
    \item \textbf{Velocity commands} $c \in \mathbb{R}^3$: target linear velocity in $x$ and $y$ directions and angular velocity about the $z$ axis.
    \item \textbf{Phase encoding} $\phi \in \mathbb{R}^2$: sinusoidal encoding $[\sin(2\pi \phi_t), \cos(2\pi \phi_t)]$ of the current gait phase.
    \item \textbf{Base angular velocity} $\omega_b \in \mathbb{R}^3$: expressed in the body frame, scaled by 0.25.
    \item \textbf{Projected gravity vector} $g_p \in \mathbb{R}^3$: the world-frame gravity vector $[0, 0, -1]^\top$ rotated by the base orientation.
    \item \textbf{Joint position error} $q_e \in \mathbb{R}^{12}$: the difference between current joint positions and the default standing configuration.
    \item \textbf{Joint velocities} $\dot{q} \in \mathbb{R}^{12}$: scaled by 0.05.
    \item \textbf{Previous action} $a_{t-1} \in \mathbb{R}^{12}$: the action executed at the previous timestep.
\end{itemize}

The phase variable $\phi_t \in [0, 1)$ is advanced as $\phi_{t+1} = (\phi_t + \Delta t / T_{\text{cycle}}) \bmod 1$, where $T_{\text{cycle}}$ is the gait-specific cycle time.

The reference joint trajectory is encoded through the phase variable. For periodic gaits (walking, goose-stepping, running, stair climbing), the reference is sinusoidal:

\begin{align}
  s &= \sin(2\pi \phi_t) \\
  q_{\text{ref}}^{\text{L,hip}} &= -s \cdot \mathbb{1}_{s<0} \\
  q_{\text{ref}}^{\text{L,knee}} &= 2s \cdot \sigma \cdot \mathbb{1}_{s<0}
  \end{align}

where $\sigma$ is the target joint position scale (0.26 for walking and running, 0.28 for goose-stepping), and the right leg mirrors the left with opposite sign when $s > 0$. Each gait modulates this pattern through its $(T_{\text{cycle}}, \text{stance\_ratio})$ pair: walking (0.8 s, 0.6), goose-stepping (0.7 s, 0.45), running (0.4 s, 0.35), stair climbing (0.7 s, 0.6).

For the jumping gait, a piecewise 4.0 s cycle trajectory is used instead, with five phases: squat ($\phi < 0.30$, $\sin^2$ knee flexion), takeoff ($0.30$--$0.42$, rapid extension), flight ($0.42$--$0.48$, hold extension), landing ($0.48$--$0.75$, impact absorption and recovery), and stand ($\phi \geq 0.75$, default pose). A curriculum progressively increases squat depth over 2000--10000 PPO iterations.

To capture temporal dynamics, the policy receives a history of $N=21$ consecutive observation frames, stacked into a single input of dimension $50 \times 21 = 1050$. This enables implicit estimation of velocities and accelerations without explicit derivative computation.

The critic receives an extended observation $s_t$ that additionally includes privileged simulation-only information: base linear and angular velocity, projected gravity, commands, joint position error, joint velocities, actions, phase encoding, friction coefficients, foot contact masks, stance masks, joint tracking error, push forces, foot vertical velocities, and foot clearance. This privileged vector (73 dimensions per frame) is stacked over 5 frames and concatenated with terrain height measurements (187 points), yielding 552 total dimensions.

\subsubsection{Action Space}

The policy outputs a 12-dimensional action vector $a_t \in \mathbb{R}^{12}$, corresponding to target joint position offsets for the 12 actuated DOFs (6 per leg: hip pitch, hip roll, hip yaw, knee, ankle pitch, ankle roll). The target joint position is computed as:

\begin{equation}
q_{\text{target}} = a_t \cdot k_{\text{action}} \cdot k_{\text{gain}} + q_{\text{default}}
\end{equation}

where $k_{\text{action}}$ is the action scale factor (0.25 for walking/stair climbing, 0.3 for running, 0.4 for goose-stepping) and $k_{\text{gain}} = \tau_{\text{max}} / K_P$ is a per-joint gain that normalizes the action output relative to the joint's stiffness and torque limit.

The target positions are sent to a PD controller:

\begin{equation}
\tau = K_P (q_{\text{target}} - q) - K_D \dot{q}
\end{equation}

The proportional gains $K_P$ range from 75 to 280 N$\cdot$m/rad for walking-type gaits (hip pitch and knee at 280, hip roll at 240, hip yaw at 75, ankle joints at 75), with derivative gains $K_D$ from 5 to 14 N$\cdot$m$\cdot$s/rad. Goose-stepping employs significantly higher stiffness (hip pitch 450, knee 800, hip roll 300, hip yaw 150, ankle joints 200) and damping ($K_D$ up to 40 on the knee), while running uses moderate increases (hip pitch and knee at 300, hip roll 240, hip yaw 100, ankle joints 80). Computed torques are clipped to per-joint maximum limits (60--150 N$\cdot$m) specified in the URDF effort field. The policy executes at 50 Hz while the PD controller runs at 200 Hz (decimation factor of 4). An exponential moving average filter smooths actions, and a random action delay of 5--10 timesteps simulates real-world latency.

\subsubsection{Reward Function}

All five gaits share a unified reward formulation. The total reward at each timestep is a weighted sum:

\begin{equation}
r_{\text{total}} = \sum_{i} w_i r_i
\end{equation}

The reward terms are organized into four categories:

\textbf{(1) Joint tracking reward.} The primary objective tracks the reference trajectory. For periodic gaits, the tracking reward for hip pitch and knee joints is computed as a dual-exponential sum:
\begin{equation}
r_{\text{track}} = 0.15 \sum_{j \in \{\text{hip,knee}\}} \sum_{i \in \{\text{L,R}\}} \left(e^{-4 e_{ij}^2} + e^{-20 e_{ij}^2}\right)
\end{equation}
where $e_{ij} = q_{ij} - q_{\text{ref},ij}$ is the joint error, with the knee term masked to apply only during swing phase. Walking and running use a general tracking weight of $w_{\text{track}} = 2.0$, goose-stepping uses $4.0$ (with tighter $\sigma = 0.15$ vs. $0.2$), and a separate knee-only reward adds $w = 3.0$ (walking) or $25.0$ (goose-stepping). For jumping, a single tracking reward ($w = 20.0$) computes mean squared error across all 12 joints.

\textbf{(2) Base balance and posture reward.} Includes orientation reward ($w = 3$ for walking and running, $5.0$ for goose-stepping, $15.0$ for jumping), base height reward ($w = 2$), and velocity tracking for linear ($w = 5.0$ walking, $8.0$ goose-stepping, $20.0$ running, $0.0$ jumping) and angular ($w = 3.5$) velocity.

\textbf{(3) Motion smoothness reward.} Action smoothness penalizes first- and second-order changes ($w = -0.01$ for walking and running, $-0.001$ for goose-stepping), joint velocity ($w = -10^{-4}$), and rated torque ($w = -0.02$ per side for walking, $-0.03$ for goose-stepping, $-0.05$ for running). Goose-stepping adds leg straightness ($w = 12.0$), calf lift ($w = 8.0$), and foot kick ($w = 6.0$) terms.

\textbf{(4) Safety and gait-specific rewards.} Fall termination when the base contacts ground or $|\text{roll}| > 0.8$ rad, $|\text{pitch}| > 1.0$ rad. Knee collision penalties are applied. Periodic gaits include feet swing height ($w = 2$) and alternate foot swing ($w = 6.0$ for goose-stepping). Jumping uses specialized rewards: jump height targeting ($w = 150.0$, Gaussian reward centered at 0.275 m with a standing-phase penalty to prevent spurious small hops), vertical takeoff velocity ($w = 120.0$, piecewise nonlinear: linear ramp below 1.0 m/s, quadratic above), and synchronized feet contact ($w = 20.0$), while penalizing horizontal velocity ($w = -15.0$).

The key design principle is that \textit{all five gaits use the same reward formulation}---diversity arises from reference specification, weight tuning, and gait-specific terms rather than structural differences.
\subsection{Selective AMP Strategy}

\subsubsection{Brief Review of AMP}

Adversarial Motion Prior (AMP) combines adversarial imitation learning with reinforcement learning to produce motions that resemble a set of reference demonstrations. The method introduces a discriminator network $D_\phi$ trained to distinguish between state transitions from expert motion clips and state transitions produced by the policy during execution.

The discriminator architecture is a 3-layer MLP with hidden dimensions $[1024, 512, 256]$ and ReLU activations. The input to the discriminator is the concatenation of the current state $s_t$ and the next state $s_{t+1}$, capturing a single state transition. The discriminator outputs a scalar logit $D(s_t, s_{t+1})$, which is converted into a reward signal:

\begin{equation}
r_{\text{AMP}} = \alpha \cdot \max\left(0, 1 - \frac{1}{4}(D(s_t, s_{t+1}) - 1)^2\right)
\end{equation}

where $\alpha$ is a per-gait coefficient scaling the AMP reward magnitude. This reward is then linearly interpolated with the task reward:

\begin{equation}
r_{\text{combined}} = (1 - \beta) \cdot r_{\text{AMP}} + \beta \cdot r_{\text{task}}
\end{equation}

where $\beta \in [0, 1]$ controls the balance between AMP regularization and task optimization.

The discriminator is trained with a binary cross-entropy loss combined with a gradient penalty for stability:

\begin{equation}
\mathcal{L}_D = \mathbb{E}_{\text{policy}}[(D + 1)^2] + \mathbb{E}_{\text{expert}}[(D - 1)^2] + \lambda \cdot \mathbb{E}_{\text{expert}}[\|\nabla_{(s,s')} D\|^2]
\end{equation}

where the gradient penalty coefficient $\lambda = 10$. The reference motion data consists of captured human locomotion sequences, processed into 200,000 preloaded state transitions.

\subsubsection{Our Selection Criterion}

The central insight of our approach is that AMP's regularization effect is beneficial for some gaits but detrimental for others. We propose a per-gait selection criterion based on motion characteristics:

\textbf{AMP is applied to:} Walking, goose-stepping, and stair climbing. These gaits are periodic with well-defined, repeating motion cycles and moderate joint amplitudes, where stability is the primary concern. For these gaits, the reference motion distribution closely matches the desired policy output, and the adversarial reward provides a stabilizing inductive bias that accelerates convergence, suppresses erratic behavior, and produces smoother joint trajectories compatible with real actuator dynamics. The AMP coefficient $\alpha$ and interpolation $\beta$ are tuned per gait: walking and stair climbing use $\alpha = 0.3$, $\beta = 0.8$; goose-stepping uses $\alpha = 0.6$, $\beta = 0.7$ (higher $\alpha$ to enforce the distinctive straight-leg style; lower $\beta$ to allow more AMP influence).

\textbf{AMP is NOT applied to:} Running and jumping. These gaits involve large joint excursions that may exceed the range of available reference data, and require rapid acceleration and explosive dynamics. For running, setting $\alpha = 0.0$ and $\beta = 1.0$ removes AMP entirely, giving the policy full freedom to discover high-speed strategies using the actuator's complete torque capacity (up to 150 N$\cdot$m on knee joints). Similarly, jumping uses $\alpha = 0.0$, $\beta = 1.0$, and disables the symmetry loss ($\text{sym\_loss} = \text{False}$) to allow asymmetric force generation during takeoff. In both cases, the AMP discriminator would penalize the policy for deviating from the reference distribution, actively discouraging the motion extremes needed for sufficient speed or jump height.

In summary: \textit{AMP improves stability but limits expressiveness; we select its use adaptively based on gait characteristics.} Table~\ref{tab:amp_config} summarizes the configuration for each gait.

\begin{table}[htbp]
\centering
\caption{Selective AMP configuration per gait.}
\label{tab:amp_config}
\begin{tabular}{lccc}
\hline
\textbf{Gait} & $\alpha$ & $\beta$ & \textbf{Sym Loss} \\
\hline
Walking & 0.3 & 0.8 & True \\
Goose-stepping & 0.6 & 0.7 & True \\
Running & 0.0 & 1.0 & True \\
Stair climbing & 0.3 & 0.8 & True \\
Jumping & 0.0 & 1.0 & False \\
\hline
\end{tabular}
\end{table}
\subsection{Training Pipeline}

\subsubsection{Simulation Training}

All policies are trained using the Proximal Policy Optimization (PPO) algorithm~\cite{c9} in the Isaac Gym physics simulator~\cite{c5} with 4096 parallel environments. The actor network uses an MLP with hidden layers [512, 256, 128] and ELU activation. The critic network shares the same architecture. The PPO algorithm runs 5 learning epochs over 4 mini-batches per iteration, with a learning rate of $5 \times 10^{-4}$, discount factor $\gamma = 0.99$, and GAE parameter $\lambda = 0.95$. The KL divergence target is set to 0.01 with an adaptive learning rate schedule. Training proceeds for up to 40,000 iterations, with the empirical observation normalizer updated online during training~\cite{c10}.

\subsubsection{Domain Randomization}

To bridge the simulation-to-reality gap, we apply extensive domain randomization during training. The following parameters are randomized at the environment level each episode:

\begin{itemize}
    \item \textbf{Contact properties}: Static and dynamic friction uniformly sampled from $[0.2, 1.5]$; restitution from $[0.0, 1.0]$.
    \item \textbf{Base mass}: Perturbed by $[-2, +5]$ kg from the nominal mass.
    \item \textbf{Link inertia}: Scaled by $[0.5, 1.8]$ independently per link.
    \item \textbf{Center of mass}: Displaced by $[-0.1, 0.1]$ m in each axis.
    \item \textbf{Motor strength}: Randomized to $[0.7, 1.0]$ (walking/stair climbing), $[0.8, 1.2]$ (running/goose-stepping), or $[0.9, 1.1]$ (jumping).
    \item \textbf{PD gains}: $K_P$ scaled by $[0.8, 1.2]$, $K_D$ scaled by $[0.8, 1.2]$.
    \item \textbf{Joint properties}: Joint friction $[0.3, 1.5]$, damping $[0.3, 4.0]$, armature $[0.8, 1.2]$.
    \item \textbf{Action lag}: Actions delayed by a random number of timesteps from $[5, 10]$, simulating control latency.
    \item \textbf{External disturbances}: Random impulses applied to the base with maximum linear velocity 2.0 m/s and angular velocity 1.5 rad/s, at intervals of 10 seconds.
    \item \textbf{Initial state}: Joint positions scaled by $[0.5, 1.5]$ and offset by $[-0.1, 0.1]$ rad at reset.
\end{itemize}

A curriculum is employed for action lag: the minimum delay is gradually increased from 2 to 5 timesteps over the course of training, starting from a milder setting to facilitate initial learning and progressively increasing the difficulty.

\subsubsection{Sim-to-Real Transfer}

Trained policies are deployed on the physical humanoid robot using zero-shot transfer---no fine-tuning on real hardware is performed. The deployment pipeline consists of:

\begin{enumerate}
    \item \textbf{Policy export}: The trained actor network is exported to ONNX format with the empirical observation normalizer as a single inference module, and loaded via ONNX Runtime on the robot's onboard computer.
    \item \textbf{Onboard execution}: The policy runs onboard the robot's computer at 50 Hz, receiving proprioceptive observations from an external IMU (angular velocity and orientation) and joint encoders (position and velocity). Since base linear velocity is not directly measurable on the hardware, the corresponding observation channels are set to zero, relying on the observation history stack to implicitly capture velocity information.
    \item \textbf{PD control}: Joint torques are computed by the onboard PD controller at 500 Hz using the target positions output by the policy. A first-order low-pass filter with cutoff frequency 50 Hz smooths the policy actions before conversion to joint targets.
\end{enumerate}

The domain randomization applied during training ensures that the policy is robust to the discrepancies between simulated and real dynamics, including differences in motor response, contact behavior, and sensor noise.

\section{Experiments}

The video of the experiments can be found at \url{https://linqi-ye.github.io/video/lite.mp4}

\subsection{Experimental Setup}

Experiments were conducted on the Lite\_11 humanoid lower-body platform, which provides 12 actuated joints spanning the bilateral hip, knee, and ankle degrees of freedom. All policies were trained in Isaac Gym and then deployed on the physical robot for validation. We considered five locomotion modes: walking, goose-stepping, stair climbing, running, and jumping. Stair climbing includes both ascent and descent, whereas jumping consists of in-place two-foot takeoff followed by landing recovery. All tasks share the same policy interface: a 12-dimensional action vector, a 50-dimensional per-frame observation, and a 21-frame history stack, resulting in a 1050-dimensional policy input. Consequently, gait diversity is induced by reference trajectories, reward allocation, and motion-prior configuration rather than by changing the control architecture.

All tasks use a unified simulation and control backbone. The physics step is 0.005~s, and the policy is updated every four simulation steps, yielding an effective control period of 0.02~s. Low-level execution is handled by a PD position controller, which converts the policy-produced joint target offsets into actuation commands. PPO is used for all experiments. Except for jumping, which uses a slightly higher entropy coefficient to accommodate stronger exploration demands, the same base hyperparameters are shared across tasks. The main training settings are summarized in Table~\ref{tab:train_setup}.

Task environments and prior settings are defined according to gait characteristics. Walking, goose-stepping, and running are trained primarily on flat or low-complexity terrain, stair climbing is trained on triangle-mesh stairs with curriculum learning, and jumping is trained on flat terrain to emphasize vertical impulse generation, aerial stabilization, and landing recovery. Selective AMP is central to the experimental design: walking, goose-stepping, and stair climbing belong to the AMP group, whereas running and jumping are trained without AMP. This partition reflects the observation that periodic, stability-oriented motions benefit from motion priors, whereas highly dynamic motions depend more strongly on expressive freedom and task-driven optimization. To improve sim-to-real transfer robustness, domain randomization is applied throughout training, including friction, added mass, center-of-mass perturbation, motor strength, PD gains, inertia, joint friction, joint damping, initial joint scaling and bias, and action delay. Stair climbing uses a stronger terrain curriculum, while jumping adopts a more conservative randomization range.

\begin{table*}[t]
\centering
\caption{Main training configurations for the five gait tasks.}
\label{tab:train_setup}
\resizebox{\textwidth}{!}{%
\begin{tabular}{lcccccccc}
\hline
\textbf{Gait} & \textbf{\# Envs} & \textbf{Obs. Dim.} & \textbf{Act. Dim.} & \textbf{Sim Step} & \textbf{Control Period} & \texttt{num\_steps\_per\_env} & \textbf{Max Iterations} & \textbf{AMP Group} \\
\hline
Walking & 4096 & 1050 & 12 & 0.005 s & 0.02 s & 24 & 40001 & AMP \\
Goose-stepping & 4096 & 1050 & 12 & 0.005 s & 0.02 s & 24 & 20001 & AMP \\
Stair climbing & 4096 & 1050 & 12 & 0.005 s & 0.02 s & 24 & 40001 & AMP \\
Running & 2048 & 1050 & 12 & 0.005 s & 0.02 s & 24 & 40001 & No AMP \\
Jumping & 2048 & 1050 & 12 & 0.005 s & 0.02 s & 24 & 11000 & No AMP \\
\hline
\end{tabular}%
}
\end{table*}

\subsection{Evaluation Metrics}

To evaluate multi-gait policies from both optimization and execution perspectives, we consider joint tracking quality, convergence speed, task reliability, training stability, and posture stability. Performance is not judged solely by whether a motion emerges, but by whether the learned motion adheres to the intended structure, converges stably during training, and remains controllable during execution.

The joint tracking metric characterizes how well the policy preserves the intended kinematic organization of each gait. For walking, goose-stepping, and stair climbing, it primarily reflects coordination between stance and swing phases; for running and jumping, it reflects the ability to maintain the target motion pattern under high-speed or explosive dynamics. Convergence steps quantify the optimization cost required to reach a stable regime. Tasks that stabilize within fewer iterations are easier to optimize under the unified formulation, whereas tasks that require more iterations place stronger demands on contact switching, posture recovery, or temporal coordination.

At execution time, success rate and fall rate describe task reliability. A high success rate indicates that the robot can repeatedly sustain the desired gait, whereas a high fall rate typically indicates sensitivity to initial conditions, rhythm control, or dynamic recovery. Training error characterizes the smoothness of optimization: lower error generally corresponds to more stable parameter updates, whereas persistently large error suggests stronger oscillation and higher uncertainty. Posture stability measures whole-body control quality. Higher posture stability corresponds to smaller torso oscillation, smoother roll-pitch regulation, and more consistent phase transitions. These metrics are used below to compare both overall gait performance and the distinction between AMP and non-AMP training.

\subsection{Results and Analysis}

\subsubsection{Overall Performance of Five Gaits}

Under a unified policy structure, shared observation design, and common reward organization, the proposed framework learns and deploys five locomotion modes: walking, goose-stepping, stair climbing, running, and jumping. These tasks span the spectrum from low-dynamic periodic locomotion to highly dynamic explosive motion, thereby testing the scalability of the unified RL formulation across substantially different gait classes.

\begin{figure}[t]
\centering
\includegraphics[width=0.85\linewidth]{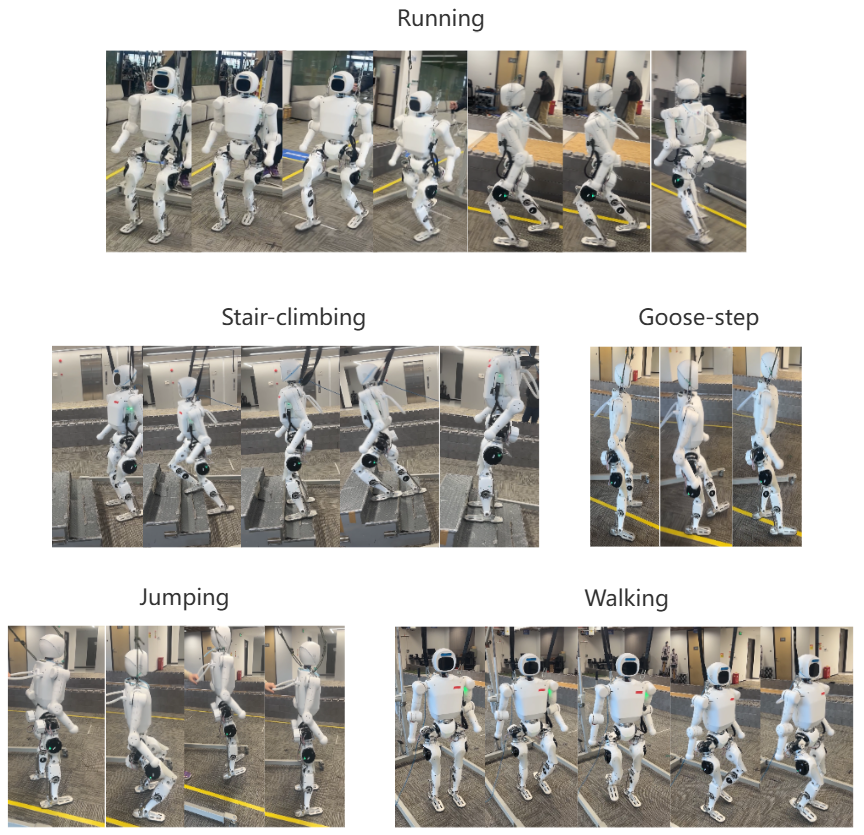}
\caption{Representative real-robot image sequences for the five learned gaits: walking, stair climbing, goose-stepping, jumping, and running.}
\label{fig:real_sequences}
\end{figure}

Qualitatively, walking yields a regular forward stepping pattern; goose-stepping exhibits the intended high leg lift and straight-knee swing; stair climbing maintains sufficient foot clearance and body posture during both ascent and descent; running produces shorter stance durations and stronger forward propulsion; and jumping realizes the full motion chain from squat preparation and rapid takeoff to aerial stabilization and landing recovery. Figure~\ref{fig:real_sequences} presents representative real-robot image sequences for the five gaits. These results indicate that the unified framework is able not only to maintain balance, but also to generate motion patterns with clearly differentiated dynamic characteristics and stylistic features.

Table~\ref{tab:gait_summary} summarizes the quantitative performance of the five gaits. The reported indicators correspond to joint tracking quality, convergence speed, task reliability, training stability, and posture stability. Because the reward composition and task difficulty differ across gaits, these values are intended to characterize training and execution behavior under the unified framework rather than to serve as strictly normalized physical quantities across tasks.

\begin{table*}[t]
\centering
\caption{Quantitative summary of the five gait tasks.}
\label{tab:gait_summary}
\resizebox{\textwidth}{!}{%
\begin{tabular}{lcccccc}
\hline
\textbf{Gait} & \textbf{Joint-Tracking Metric} & \textbf{Convergence Steps} & \textbf{Estimated Success Rate} & \textbf{Training Error} & \textbf{Posture-Stability Metric} & \textbf{Motion Characteristic} \\
\hline
Walking & $\ 0.40$ & $\ 2500$ & $\ 95\%$ & $\ 10.0$ & $\ 2.3$ & Periodic and stable forward locomotion \\
Goose-stepping & $\ 0.94$ & $\ 2500$ & $\ 97\%$ & $\ 5.4$ & $\ 1.2$ & High leg lift and straight-knee swing \\
Stair climbing & $\ 1.00$ & $\ 5200$ & $\ 85\%$ & $\ 1.0$ & $\ 0.95$ & Stair ascent/descent with recovery \\
Running & $\ 0.50$ & $\ 3000$ & $\ 96\%$ & $\ 4.8$ & $\ 1.3$ & Short stance and strong propulsion \\
Jumping & $\ 15.5$ & $\ 3000$ & $\ 83\%$ & $\ 6.0$ & $\ 8.6$ & Two-foot takeoff and landing recovery \\
\hline
\end{tabular}%
}
\end{table*}

\subsubsection{Effect of AMP on Different Gaits}

This subsection examines the operating boundary of AMP across gait classes. In general, AMP is most effective for gaits with clear periodic structure and stability-first objectives, whereas it can become restrictive for highly dynamic and strongly explosive motions. Walking, goose-stepping, and stair climbing fall into the former category: they require stable stance--swing transitions, consistent joint coordination, and low postural fluctuation. For such tasks, motion priors reduce ineffective exploration and typically yield lower tracking error, smoother reward growth, and faster convergence.

\begin{figure*}[t]
\centering
\includegraphics[width=0.98\textwidth]{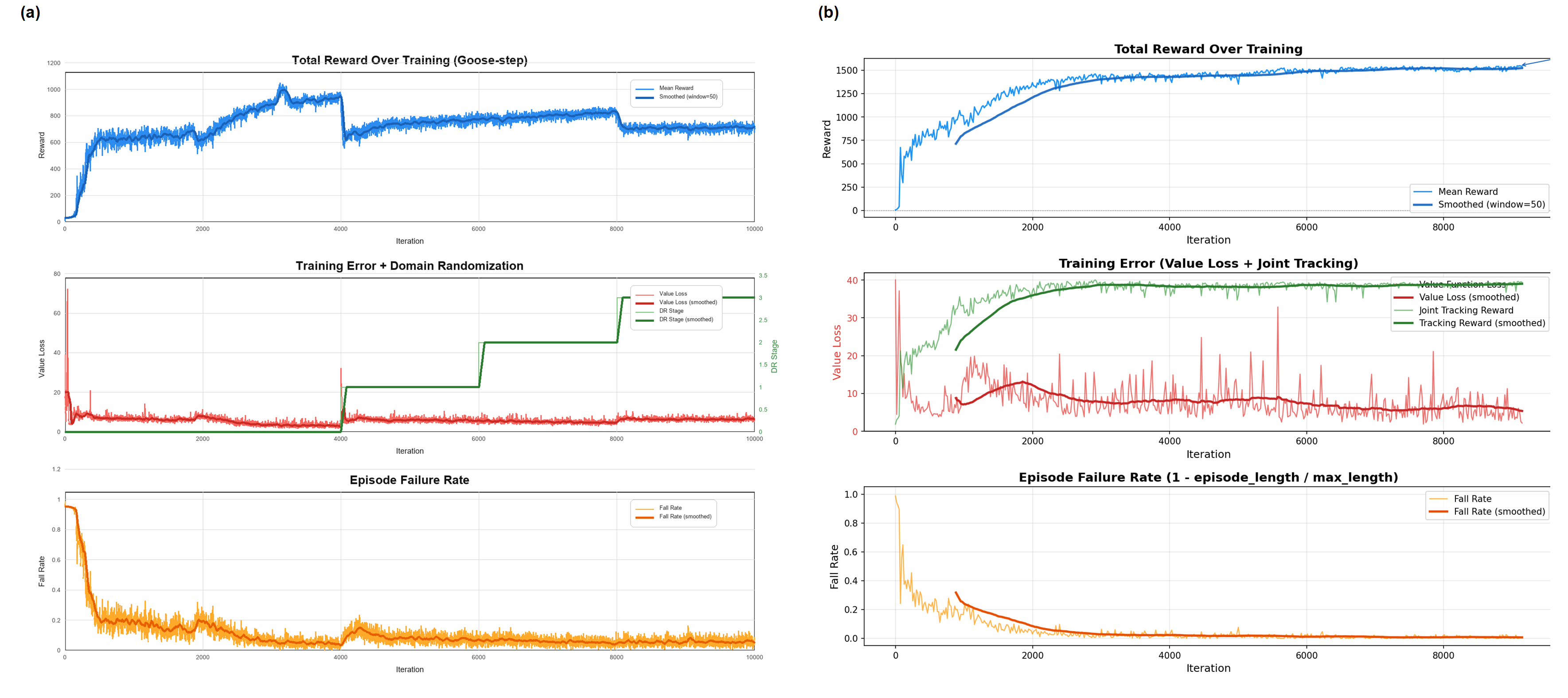}
\caption{Training curves used in the AMP-versus-no-AMP comparison. (a) Goose-stepping with AMP. (b) Jumping without AMP. Each panel reports total reward, training error or auxiliary metric, and failure rate.}
\label{fig:amp_vs_noamp}
\end{figure*}

Figure~\ref{fig:amp_vs_noamp} compares goose-stepping with AMP and jumping without AMP as representative cases. Under AMP, goose-stepping exhibits a rapid increase in total reward, low training error, and a prompt decay in failure rate. The main contribution of AMP in this setting is not merely a higher terminal return, but the reinforcement of periodic structure and style consistency, which is especially important for a gait with explicit kinematic features such as high leg lift and straight-knee swing. For running, a strong AMP prior would not necessarily render the policy incapable of locomotion, but it would likely increase behavioral conservatism and bias the solution toward a more strongly regularized motion pattern. In that regime, the learned behavior would be more likely to resemble a smooth fast walk or a weak running pattern, rather than a genuinely explosive gait with clear aerial phases, strong forward propulsion, and large leg excursion. This observation suggests that, for highly dynamic locomotion, an excessively strong motion prior may improve motion regularity while simultaneously compressing the policy's access to high-amplitude dynamic solutions. By contrast, jumping depends on large motion amplitude, vertical impulse, and aerial freedom. Under the non-AMP setting, the task still attains sustained reward growth and low failure rate while the joint-tracking signal improves steadily. This indicates that task rewards alone are sufficient to drive effective takeoff, flight, and landing recovery when expressive exploration is preserved. These results support the view that AMP should be treated as a gait-dependent mechanism rather than a uniformly active module.

\subsubsection{Real-Robot Robustness and Practical Performance}

Beyond single demonstrations, the engineering value of the method is reflected in sustained execution on the physical robot. We therefore assess real-robot robustness from three perspectives: long-horizon continuity, consistency across repeated trials, and recovery under mild external disturbances or start--stop transitions.

The robustness criteria vary across gaits. For walking and goose-stepping, the emphasis lies in rhythm consistency and controlled torso oscillation; for stair climbing, in foot clearance, step placement, and posture recovery after stair transitions; for running, in stability during rapid short-stance switching; and for jumping, in synchronized takeoff, aerial posture control, landing absorption, and return to the next motion cycle. In particular, the jumping reference is explicitly divided into squat, takeoff, flight, landing, and standing-recovery phases, so robustness depends not only on leaving the ground, but also on the quality of aerial motion and landing recovery.

Taken together, the real-robot deployments show that the learned policies are not limited to visually plausible motions in simulation, but exhibit the continuity and repeatability required for physical execution. This supports the practical viability of the proposed unified framework for multi-gait humanoid locomotion.

\section{Conclusion}

This paper presented a unified reinforcement learning framework for multi-gait humanoid locomotion together with a selective AMP strategy. Within a common policy architecture and reward formulation, the proposed method learned five representative gaits, namely walking, goose-stepping, stair climbing, running, and jumping, and transferred them to a physical humanoid platform. The results indicate that motion priors are beneficial for periodic, stability-oriented gaits, but should be weakened or removed for highly dynamic behaviors that depend on expressive motion generation. More importantly, the proposed framework provides a practical and reproducible baseline for multi-gait learning on real humanoid hardware by combining a unified RL formulation, gait-dependent prior selection, and real-robot validation. Overall, this study suggests that practical humanoid mobility will depend not only on stronger policies, but also on principled methods for organizing diverse behaviors within a shared learning framework.



\begin{thebibliography}{99}

\bibitem{c1} L. Bao, L. Humphreys, and J.~C.~G. Pimentel, ``Deep
reinforcement learning for robotic bipedal locomotion: A brief survey,''
\emph{arXiv preprint arXiv:2404.17070}, 2024.

\bibitem{c2} V. Makoviychuk, L. Wawrzyniak, Y. Guo, M. Liu, K. Mack, M. Mack,
G. State, B. Sundaralingam, Y. Zhu, and Y. Xian, ``Isaac Gym: High performance
GPU-based physics simulation for robot learning,'' in \emph{Proc. NeurIPS
Datasets and Benchmarks}, 2021.

\bibitem{c3} X. Gu, Y.-J. Wang, and J. Chen, ``Humanoid-Gym: Reinforcement
learning for humanoid robot with zero-shot sim2real transfer,'' \emph{arXiv
preprint arXiv:2404.05695}, 2024.

\bibitem{c4} I. Radosavovic, T. Xiao, B. Zhang, T. Darrell, J. Malik, and
K. Sreenath, ``Real-world humanoid locomotion with reinforcement learning,''
\emph{Sci. Robot.}, vol.~9, no.~89, p.~eadi9579, 2024.

\bibitem{c5} L. Bao, L. Humphreys, and J.~C.~G. Pimentel, ``Gait-conditioned
reinforcement learning with multi-phase curriculum for humanoid locomotion,''
in \emph{Proc. IEEE-RAS Int. Conf. Humanoid Robots (Humanoids)}, 2024.

\bibitem{c6} X.~B. Peng, Z. Ma, P. Abbeel, S. Levine, and A. Kanazawa, ``AMP:
Adversarial motion priors for stylized physics-based character control,''
\emph{ACM Trans. Graph. (SIGGRAPH)}, vol.~40, no.~4, pp.~1--20, 2021.

\bibitem{c7} A. Escontrela, X.~B. Peng, W. Yu, T. Zhang, A. I\c{s}\c{c}en,
K. Goldberg, and P. Abbeel, ``Adversarial motion priors make good substitutes
for complex reward functions,'' in \emph{Proc. IEEE/RSJ Int. Conf. Intell.
Robots Syst. (IROS)}, 2022, pp.~223--230.

\bibitem{c8} F. Lerario, G. Nava, A. Loquercio, and D. Scaramuzza, ``Learning
to walk and fly with adversarial motion priors,'' in \emph{Proc. IEEE/RSJ Int.
Conf. Intell. Robots Syst. (IROS)}, 2024, pp.~8976--8983.

\bibitem{c9} J. Schulman, F. Wolski, P. Dhariwal, A. Radford, and O. Klimov,
``Proximal policy optimization algorithms,'' \emph{arXiv preprint
arXiv:1707.06347}, 2017.

\bibitem{c10} X.~B. Peng, M. Andrychowicz, W. Zaremba, and P. Abbeel,
``Sim-to-real transfer of robotic control with dynamics randomization,'' in
\emph{Proc. IEEE Int. Conf. Robot. Autom. (ICRA)}, 2018, pp.~3567--3574.

\end{thebibliography}
\end{document}